\documentclass{IEEEtran4PSCC}

\usepackage{cite}

\ifCLASSINFOpdf
   \usepackage[pdftex]{graphicx}
\else
   \usepackage[dvips]{graphicx}
\fi

\usepackage[cmex10]{amsmath}
\usepackage[english]{babel}
\usepackage[dvipsnames]{xcolor}
\usepackage{amssymb, amsthm}
\usepackage[boxed]{algorithm2e}
\usepackage{graphicx} 
\usepackage{placeins}
\usepackage{subcaption} 
\usepackage{amsfonts}
\usepackage{enumerate,enumitem}
\usepackage{exercise}
\usepackage{ragged2e}
\usepackage{caption}
\usepackage{tikz}
\usepackage{pgfplots}
    \usepgfplotslibrary{fillbetween}
\usepackage{mathtools}
\usepackage{microtype}
\usepackage{hyperref}
\usepackage[nameinlink, capitalise]{cleveref}

\newcommand{\A}{\mathcal{A}}
\newcommand{\mS}{\mathcal{S}}

\newcommand{\R}{\mathbb{R}}

\newcommand{\E}{\mathbb{E}}
\usepackage{bbm}

\newcommand{\bbm}{\begin{bmatrix}}
\newcommand{\ebm}{\end{bmatrix}}

\makeatletter
\let\old@ps@headings\ps@headings
\let\old@ps@IEEEtitlepagestyle\ps@IEEEtitlepagestyle
\def\psccfooter#1{%
    \def\ps@headings{%
        \old@ps@headings%
        \def\@oddfoot{\strut\hfill#1\hfill\strut}%
        \def\@evenfoot{\strut\hfill#1\hfill\strut}%
    }%
    \def\ps@IEEEtitlepagestyle{%
        \old@ps@IEEEtitlepagestyle%
        \def\@oddfoot{\strut\hfill#1\hfill\strut}%
        \def\@evenfoot{\strut\hfill#1\hfill\strut}%
    }%
    \ps@headings%
}
\makeatother

\psccfooter{%
        \parbox{\textwidth}{\hrulefill \\ \small{23rd Power Systems Computation Conference} \hfill \begin{minipage}{0.2\textwidth}\centering \vspace*{4pt} \includegraphics[scale=0.06]{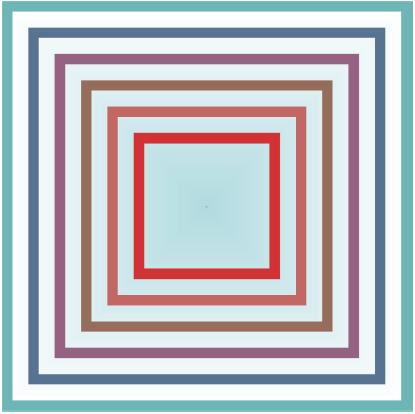}\\\small{PSCC 2024} \end{minipage} \hfill \small{Paris, France --- June 4 -- 7, 2024}}%
}

\begin{document}
\title{Multi-Agent Reinforcement Learning for\\Power Grid Topology Optimization}

\author{
\IEEEauthorblockN{Erica van der Sar, Alessandro Zocca, Sandjai Bhulai}
\IEEEauthorblockA{Department of Mathematics,
Vrije Universiteit,
Amsterdam, The Netherlands\\
\{e.t.van.der.sar, a.zocca, s.bhulai\}@vu.nl}
}

\maketitle

\begin{abstract}
Recent challenges in operating power networks arise from increasing energy demands and unpredictable renewable sources like wind and solar. While reinforcement learning (RL) shows promise in managing these networks, through topological actions like bus and line switching, efficiently handling large action spaces as networks grow is crucial. This paper presents a hierarchical multi-agent reinforcement learning (MARL) framework tailored for these expansive action spaces, leveraging the power grid's inherent hierarchical nature. Experimental results indicate the MARL framework's competitive performance with single-agent RL methods. We also compare different RL algorithms for lower-level agents alongside different policies for higher-order agents.
\end{abstract}


\section{Introduction}
In 2019, the French TSO RTE launched the \emph{Learning to Run a Power Network} (L2RPN) challenge \cite{marot2020learning}, encouraging diverse researchers to use reinforcement learning (RL) for power network maintenance. While various solutions emerged, see \cite{yoon2021winning,zhou2021action,Subramanian2021} and more recently \cite{dorfer2022power,Matavalam2023}, all faced challenges with the vast combinatorial action space, including line-switching and bus-splitting. This underscores the need for an advanced RL framework ensuring scalability and easy integration of actions like generation redispatch.

Multi-Agent Reinforcement Learning (MARL) seems a well-suited approach to address these challenges due to their scalability. 
However, its potential in this domain remains underexplored. A recent paper by \cite{anonymous2023hierarchical} introduced a hierarchical reinforcement learning (HRL) framework where RL agents have roles based on their operational level. Still, a primary agent determines the best topological action.

A recent paper \cite{yoon2021winning} utilizes a hierarchical framework that activates a soft actor-critic (SAC) agent \cite{haarnoja2018soft} only when issues related to grid safety arise and uses graph attention layers to learn the intricate dependencies within the power network. While this SAC approach keeps a small action space due to its after-state implementation, its adaptability to network shifts is limited. To address this limitation, we transitioned to a soft actor-critic discrete (SACD) agent \cite{christodoulou2019soft} that learns directly from actions rather than the after-state. 

Although this adjustment does lead to a larger action space, we distribute tasks to multiple RL agents per substation instead of one global agent. This change brings benefits like reduced action space per agent, enhanced scalability, and easier integration of actions like generator redispatching, forming a cooperative MARL framework \cite{oroojlooy2023review}. 

The simplest \textit{independent learning} MARL treats each agent independently and considers the rest of the agents as part of the environment \cite{tan1993multi}. This approach avoids scalability and communication issues, however, the non-stationarity of the environment from each agent's perspective slows and possibly hinders learning. We use the \textit{centralized training with decentralized execution} strategy \cite{lowe2017multi}, allowing agents to access additional information during training but not during execution.

This paper introduces various MARL strategies for power grid control, utilizing the problem's inherent hierarchical structure. In addition to SACD, we use another state-of-the-art RL algorithm, proximal policy optimization (PPO) \cite{schulman2017proximal}.

The paper is organized as follows: \cref{sec:background} discusses Grid2Op and previous solutions on L2RPN; \cref{sec:methods} details our RL methods; \cref{sec:results} presents our findings; and \cref{sec:conclusion} concludes and outlines future work.

\section{Background and Related work} \label{sec:background}

\subsection{Controlling a power network: the Grid2Op environment}\label{SubSec:L2RPN_EnvG2Op}
\begin{figure}[!tbh]
    \centering
    \includegraphics[width=0.98\linewidth]{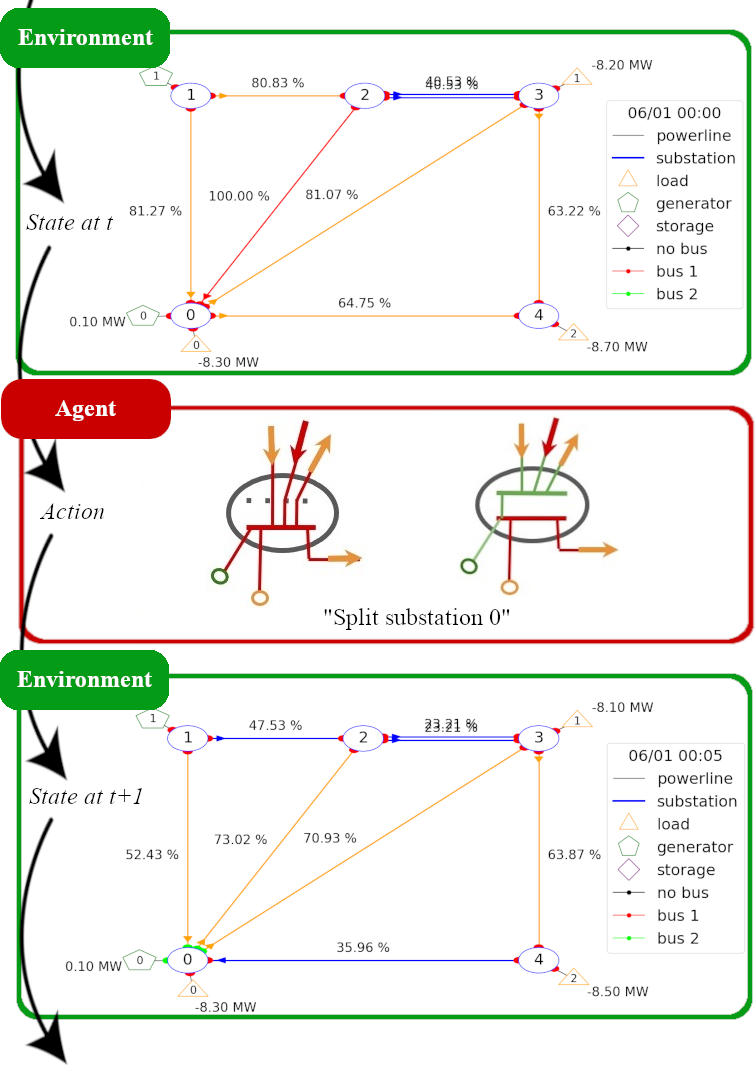}
    \caption{Example of a bus-switching action by an agent and the effect on the environment. Some elements of substation 0 are assigned to bus 2, which results in a split of substation 0, and a consequent power flow redistribution.} 
    \label{Fig:Grid2OpExpl}
\end{figure}

This section introduces Grid2Op \cite{Grid2Op}, an open-source framework used for the \emph{Learning to Run a Power Network} (L2RPN) competitions \cite{marot2020learning,Marot2021,Marot2022}. Designed for sequential decision-making in power systems, its aim is to ensure a safe power network by avoiding contingencies and ensuring constant connectivity. Grid2Op simulates real-world power grids, depicting them as graphs with nodes as substations and edges as transmission lines or transformers. Each substation connects various elements like generators, loads, and storage units, and connects to one of two \textit{buses}. A consistent bus connection makes a substation a singular node; differing connections `split' it, changing the local network topology. Grid2Op can compute the resulting power flow configurations from any topology change; see an example in \cref{Fig:Grid2OpExpl}. 
Modifications in a substation's bus configuration are termed \textit{bus-switching actions}. Grid2Op simulates other actions like \textit{line-switching} and \textit{power redispatch actions}. Cost-effective for responding to contingencies, these \textit{topological actions} redistribute line flows, mitigating issues without the need for pricier generators or load adjustments.

In this paper, we focus solely on bus-switching actions, and we use the realistic Grid2Op environment to develop a sequential decision-making RL model, termed \textit{agent}, that keeps the power grid in a safe regime by choosing optimal bus-switching actions.

\subsection{Previous solution approaches for L2RPN} \label{SubSec:OtherSolutions} 
The winning approach of L2RPN WCCI 2020 was the Semi-Markov Afterstate Actor-Critic (SMAAC) proposed by \cite{yoon2021winning}. It acts only in dangerous states, transitioning from a Markov to a semi-Markov decision process. Instead of learning specific actions, SMAAC learns the optimal bus substation configuration, termed \textit{goal topology}. Given that only one substation can be acted on at once in L2RPN, determining the action sequence is crucial. The authors of \cite{yoon2021winning} used a two-tiered system: the ``high-level" SAC-based policy sets the goal, while the ``low-level" policy defines the sequence, with four rule-based options tested for the latter.
SMAAC's goal topology ensures the SAC algorithm learns fewer values. Yet, this leads to an agent that is largely unresponsive to environmental shifts. It tends to stick to one effective configuration without adapting as needed. Small actor output changes often do not influence the action unless they surpass a set threshold, complicating course correction during gradient descent.

Another recent paper \cite{anonymous2023hierarchical} employs a three-tiered structure for optimal power network topology control. Like \cite{yoon2021winning}, the top level is rule-based. The intermediate-level agent selects the substation and utilizes PPO or SAC-trained policies. The lowest level uses either a brute-force method examining all configurations or an RL agent with an action mask aligned with the intermediate-level agent's choice.

\section{Methods}
\label{sec:methods}
In this section, we provide details about the RL environment we used and our proposed RL architectures and algorithms.

\subsection{Reinforcement learning environment: Grid2Op}
The Grid2Op environment described in \cref{SubSec:L2RPN_EnvG2Op} is typically described as a Markov Decision Process (MDP) defined by $(\mS,\A, p, r)$, where at each time step $t$ an agent observes a state $s_t\in \mS$ from the environment and takes an action $a_t \in \A$. The environment returns the next state $s_{t+1}\in \mS$ to the agent with probability $p(s_{t+1}|s_t, a_t)$, which is the unknown state transition probability, and the agent receives an immediate reward $r(s_t, a_t) \in \R$. With this MDP formulation, reinforcement learning (RL) can be used to learn a (stochastic) policy $\pi(a_t|s_t)$ that optimizes the expected discounted reward $\E_\pi [\sum_{t=0}^T \gamma^t r(s_t, a_t)]$, where $\gamma \in (0,1)$ is the discount factor. 

Within the Grid2Op environment, the state $s_t$ that the agent can observe at time $t$ consists of (i) the current generator and load states, including power production/consumption; (ii) the current topology configuration, comprehensive of line connections and current bus configuration at each substation; and (iii) the current load $\rho_{\ell} \in [0,1]$ on each line $\ell$, measured as the fraction of the capacity of that power line.

The agent must choose a bus-switching action $a_t$ from the fixed collection $\A$ of substation reconfigurations. These always include the trivial action, which we name \textit{do-nothing action}, that does not change the current bus configuration. As mentioned in \cref{sec:background}, the objective is to keep the power network in a \textit{safe} state cost-effectively throughout each episode. Congested lines auto-disconnect based on overload severity and have a reconnection delay. Any network disconnection or isolation of a load or generator results in a ``game over" with a reward of $-1$.

To steer the agent towards learning how to maintain a safe power grid, it is important to have a proper reward function that penalizes whenever power lines are in overload. When lines are congested, it leads to an increase in energy loss within the network. Therefore, during training, we use the energy efficiency of the power grid as a reward function defined by rescaling the ratio of the total served load and the total generation, similarly to \cite{yoon2021winning}. 

\subsection{Reinforcement learning algorithms} \label{sec:RLalgo}
Next, we introduce two RL algorithms: a discrete Soft Actor-Critic variant (SACD) and Proximal Policy Optimization (PPO). In both algorithms, the policy, indicated by $\pi$, will be parameterized using a neural network with learnable parameters $\theta$, while the action-value function, denoted by $Q(s_t, a_t)$, and for PPO the state-value function $V(s_t)$ will be represented by a neural network with parameters $\phi$.

\subsubsection{Soft actor-critic discrete (SACD)}
Soft Actor-Critic (SAC) \cite{haarnoja2018soft} is an off-policy reinforcement learning algorithm aiming for efficient and stable learning in continuous action spaces. The algorithm incorporates an entropy term into its reward function, effectively encouraging more exploratory behavior. This leads to improved state-space exploration and provides a degree of robustness in policy learning. The core advantage of SAC is its ability to balance exploration and exploitation efficiently. Soft actor-critic discrete (SACD) is an adjustment of SAC to make the algorithm applicable to discrete environments \cite{christodoulou2019soft}.

For the critic, we learn a soft q-function $Q_\phi(s,a)$ via off-policy temporal-difference learning by minimizing the critic-loss:
\begin{align}
    J_{Q}(\phi)&= \E_{(s_t,a_t) \sim D } \Big[ \frac{1}{2} \big( Q_ {\phi}(s_t,a_t) - y(s_t,a_t) \big)^2 \Big]  \label{Eq_SACD_critic} \\
    \textnormal{with } & y(s_t, a_t) = r_t + \gamma \E [ V_{\bar{\phi}}(s_{t+1})], \nonumber
\end{align}
where $D$ is a replay buffer of past experiences and $\bar{\phi}$ is the parameter for the target critic network, and the soft state value calculation is defined as
\begin{align*}
     V_{\bar{\phi}}(s_{t+1}) = \pi_\theta(s_{t+1})^T \big[ Q(s_{t+1}) - \alpha \log \big(\pi_\theta(s_{t+1}) \big) \big],
\end{align*}
where $\alpha$ determines the relative importance of the entropy term versus the reward and is called the temperature parameter.

The objective function of the policy, the actor-loss, is:
\begin{align}
    J_\pi(\theta) = \E_{s_t \sim D} \Big[ \pi(s_t) \big[ \alpha \log \big(\pi_\theta(s_t) \big) - Q_\phi(s_t) \big] \Big].
    \label{Eq_SACD_actor}
\end{align}

The objective function for the temperature parameter $\alpha$ is 
\begin{align*}
    J(\alpha) = \pi_\theta (s_t)^T \Big[ -\alpha \Big( \log \big(\pi_\theta (s_t) \big) + \bar{H} \Big) \Big],
\end{align*}
with $\bar{H}$ a constant vector equal to the hyperparameter representing the target entropy.

\subsubsection{Proximal Policy Optimization (PPO)}
Proximal Policy Optimization (PPO) \cite{schulman2017proximal} is a policy optimization algorithm that improves upon traditional policy gradient methods by introducing a clipped surrogate objective function. This ensures that policy updates remain close to the original policy, effectively balancing the trade-off between exploration and exploitation. For PPO, the policy loss is computed as 
\begin{align*}
    L^{\text{CLIP}}_t(\theta) &= \E \Big[ \min \Big( r_t(\theta) A_t, \text{clip} \big(r_t(\theta), 1-\epsilon, 1+\epsilon \big) A_t \Big) \Big],
\end{align*}
with $\epsilon$ a clipping hyperparameter, $r_t(\theta) = \pi_\theta(a_t | s_t) / \pi_{\bar{\theta}}(a_t | s_t)$ denotes the probability ratio with $\bar\theta$ the parameters of the current policy, and $A_t$ is the generalized advantage estimation (GAE) function defined as
\begin{align*}
    A_t &= \sum_{l=0}^h (\gamma \lambda)^l \big[ r_{t+l} + \gamma V_\phi(s_{t+l+1}) - V_\phi(s_{t+l}) \big],
\end{align*}
where $h$ is the number of time steps collected in each iteration. The critic-loss is again a squared error loss
\begin{align*}
    L^{\text{VF}}_t(\phi) = \big( V_\phi(s_t) - V_t^{\text{targ}} \big)^2 = \Big(V_\phi(s_t) - (A_t + V_{\bar{\phi}}(s_t) \big) \Big)^2.
\end{align*}

Combining both losses, the following objective is obtained, which is minimized in each iteration:
\begin{align*}
    L_t
    = \E_t \Big[ -L_t^{\text{CLIP}}(\theta) + c_1 L^{\text{VF}}_t(\phi) - c_2 S\big[\pi_\theta (s_t)\big] \Big],
\end{align*}
where $c_1,c_2$ are coefficients, and $S$ denotes an entropy bonus.

\subsection{Hierarchical architecture}\label{Sec:HierarchialStruc}
Building on suggestions from \cite{yoon2021winning} and \cite{anonymous2023hierarchical}, power network control naturally fits a hierarchical scheme due to the tiered nature of topological actions. Decisions range from determining if an action is needed, selecting substations to act upon, to finalizing their busbar configuration. We introduce a three-level hierarchical reinforcement learning framework, detailed subsequently.

\subsubsection{Highest level}
At the highest level, a single agent determines whether to act at each time step, often doing nothing in safe environments. We use a rule-based method, similar to prior L2RPN implementations. The agent checks for any critically loaded line, with a load $\rho$ exceeding a threshold $\rho_{\text{thresh}}$.  When the environment is \textit{unsafe}, that is $\max_\ell \rho_\ell > \rho_{\text{thresh}}$, the highest-level agent activates the mid-level one. It remains passive while lower agents work, evaluating environment safety after their actions conclude.

\subsubsection{Mid level}\label{Sec:MiddleAgent}
The intermediate level has a single agent that, upon activation, selects the substations requiring action. Since network operators typically prefer to intervene at a single substation at a time, the mid-level agent is also tasked with establishing the order in which the selected substations should act. Taking a rule-based approach, we apply the CAPA policy from \cite{yoon2021winning}, known for its effectiveness in the context of goal topology. This policy prioritizes substations with higher line loads. The mid-level agent then sequentially activates the low-level agents based on the chosen order.

\subsubsection{Lowest level}\label{Sec:LowestAgent} 
The lowest level features \textit{substation-specific agents}, each responsible for selecting bus assignments for their substation's elements. These agents have a predefined discrete action space, further reduced by excluding symmetric actions. Unlike rule-based higher-level agents, these agents use reinforcement learning to determine optimal policies, aligning with the Multi-Agent Reinforcement Learning (MARL) framework. Our unique approach of multiple substation-specific agents for topology decisions distinguishes our method from prior hierarchical strategies. The subsequent section delves into the MARL architectures we explore.

\subsection{Multi-agent reinforcement learning (MARL)} \label{Sec:MARL}
As mentioned in \cref{Sec:LowestAgent}, for the lowest level agent, we use a Multi-Agent Reinforcement Learning (MARL) framework where multiple agents function in a shared environment. 

We explored two MARL variants with independent and dependent agents. Independent agents each learn their own policy, viewing others as part of the environment. This makes the environment non-stationary for each agent due to others' learning, eliminating convergence assurances. However, independent agent structures have empirically shown effective performance. In contrast, dependent agents, collaborating and sharing information, can produce more coherent and effective multi-agent policies. Yet, handling inter-agent dependencies can increase computational costs and scaling difficulties. 
Hence, for the dependent agents, we embrace the Centralized Training with Decentralized Execution (CTDE) paradigm, a favored approach in cooperative MARL \cite{Lowe2017,Foerster2018}. In this paradigm, agents train using global state information for collective learning, but make decisions via decentralized local policies.

We evaluate two primary model-free RL algorithms, Proximal Policy Optimization (PPO) and the Soft Actor-Critic with a discrete action space (SACD) from \cref{sec:RLalgo}, resulting in four distinct MARL strategies:
\begin{enumerate}
    \item Independent agents SACD (ISACD); 
    \item Independent agents PPO (IPPO); 
    \item Dependent agents SACD (DSACD);
    \item Dependent agents PPO (DPPO).
\end{enumerate}
In the next subsections, we elaborate on each strategy. Regardless of the chosen approach, the mid-level agent's activation sets a sequence for the low-level agents. Although the sequence persists even if a low-level agent takes no action, we omit such actions to save time and reduce ``game over" risks.

\subsubsection{Independent SACD (ISACD)} \label{Sec:ISACD}
In the independent MARL version, each agent has its own critic and actor, which are updated based on their own state-action pairs. For each agent $i$, we update its state-action value-function $Q^i_\phi$ and its policy $\pi^i_\theta$ using \eqref{Eq_SACD_critic} and \eqref{Eq_SACD_actor} respectively, based on the transitions stored in $D^i$. Thus, to compute the critic-loss for agent $i$, we have:
\begin{align}
    J_{Q^i}(\phi) &= \E_{(s_t,a_t) \sim D^i } \Big[ \frac{1}{2} \big( Q^i_ {\phi}(s_t,a_t) - y(s_t,a_t) \big)^2 \Big]  \label{Eq:ISACD_critic}
\end{align}
with
\begin{align*}
y(s_t, a_t) &= r_t + \gamma \E \big[ V^i_{\bar{\phi}}(s_{t+1}) \big], \nonumber
\end{align*}
where the soft state-value function of the next state is calculated based on the agents' $i$ own policy and critic:
\begin{align*}
     V^i_{\bar{\phi}}(s_{t+1}) = \pi^i_\theta(s_{t+1})^T \Big[ Q^i(s_{t+1}) - \alpha \log \big( \pi^i_\theta(s_{t+1}) \big) \Big].
\end{align*}

\subsubsection{Independent PPO (IPPO)}\label{Sec:IPPO}
Similarly to the independent MARL version for SACD, in IPPO, the policy and critic-loss for each agent $i$ are calculated based on their own replay buffer $D^i$. For each agent $i$ we, will update the policy loss with 
\begin{align*}
    L^{\text{CLIP}}_t(\theta^i) &= \E \Big[ \min \Big(r_t(\theta^i) A_t^i, \text{clip} \big(r_t(\theta^i), 1-\epsilon, 1+\epsilon \big) A_t^i \Big) \Big],
\end{align*}
where $ r_t(\theta^i) = \pi_{\theta^i}(a_t | s_t) / \pi_{\bar{\theta}^i}(a_t | s_t)$ and the GAE becomes
\begin{align}
    A_t^i = \sum_{l=0}^h (\gamma \lambda)^l \delta^i_{t+l} 
    \, \, \textnormal{with} \, \, \delta^i_t = r^i_t + \gamma V^i(s_{t+1}) - V^i(s_t).\label{Eq:IPPO_GAE_delta}
\end{align}
The critic-loss and combined loss are computed per agent $i$.

\subsubsection{Dependent SACD (DSACD)}\label{Sec:DSACD}
The independent learning method, while simple, often struggles with non-stationarity and may not achieve the optimal policy due to limited information sharing during training. A proposed solution is using a centralized critic, incorporating all agents' actions $Q(s,a^1,\dots, a^n)$. However, this approach risks the curse of dimensionality since we can have $\prod_{i=1}^n |A^i|$ number of action combinations.

The L2RPN challenge environment's structure ensures agents do not act simultaneously within a time step. As detailed in \cref{Sec:MiddleAgent}, lower-level agents activate sequentially based on substation loads. There is a certain probability $p_{ij}$ that the lower-level agent $j$ acts after agent $i$, based on the load distribution in this substation. These distributions are not known a priori, so they are estimated during the training phase. These probabilities $p_{ij}$ are stored in matrix $\hat{\pi}_{\text{mid}}$, which is used to update the value functions of the critics, making the update of agent $i$ dependent on the current value function of the other agents. We still use \eqref{Eq:ISACD_critic} to compute the critic-loss for agent $i$, but modify the expression $y$ into $y(s_t, a_t) = r_t + \gamma \E [ \hat{V}(s_{t+1})]$. Here, $\hat{V}(s_{t+1})$ is the average soft state-value function accounting for the likelihood of each agent to act after agent $i$, i.e.,
\begin{align*}
    \hat{V}(s_{t+1}) = (\hat\pi^i_{\text{mid}})^T 
    \begin{bmatrix}
        V^1(s_{t+1})\\
        \vdots\\
        V^n(s_{t+1})
    \end{bmatrix},
\end{align*}
where $\hat\pi^i_{\text{mid}} = [ p_{i1} \dots p_{in}]^\top$ is a vector whose $j$-th entry is the empirical probabilities of activating agent $j$ after agent $i$ using the current mid-level agents' policy $\pi_{\text{mid}}$.

\subsubsection{Dependent PPO (DPPO)}\label{Sec:DPPO}
Like the DSACD agent, the DPPO agent will use the probability matrix $\hat{\pi}_{\text{mid}}$, but now to update the GAE. All equations remain the same, except for the temporal difference in \eqref{Eq:IPPO_GAE_delta}, which changes to  
\begin{align*}
    \delta^i_t &= r^i_t + \gamma 
    (\hat\pi^i_{\text{mid}})^T 
    \begin{bmatrix}
        V^1(s_{t+1})\\
        \vdots\\
        V^n(s_{t+1})
    \end{bmatrix}  
    - V^i(s_t)
\end{align*}
to reflect a dependent update for each agent $i$.

\section{Results} \label{sec:results}
In this section, we present the results obtained using both single-agent and multi-agent architectures in the L2RPN context.

\subsection{Experimental setup} \label{sec:setup}
We apply our RL-based design to the \textit{IEEE case 5} environment, a power grid with 5 substations, 8 lines, 2 generators, and 3 loads, shown in \cref{Fig:Grid2OpExpl}. This environment has 20 episodes, or \textit{chronics}, per Grid2Op documentation. Each episode has 2016 time steps, representing 5-minute intervals, showcasing varying demand and supply patterns. For diverse training starts, episodes are divided into five overlapping sub-episodes of 864 time steps, equaling three days. We allocated 18 episodes for training, one for testing, and one for validation. 

Each of the four RL architectures was trained for 10,000 interactions across five model seeds, counting only steps where the low-level agents were active. For every 100 interactions, agent performance was assessed using test sub-episodes and the L2RPN Codalab competition's rescaled score function. This score differs from the reward function used during training, but it captures the agents’ performance well. Using this function, the agent receives a score between $-100$ and $0$ when the agent dies before the Do-Nothing baseline agent would, a score above $0$ when it outperforms the baseline, and a score between $80$ and $100$ if the agent is able to finish the episode. 

Substations with less than three connected elements are not as crucial to control, since most actions either disconnect elements or resemble line-switching actions. We limit agents to substations of size larger than 3, resulting in three low-level agents in the MARL setup.

\begin{figure}[!tb]
    \centering
    \includegraphics[width=0.99\linewidth]{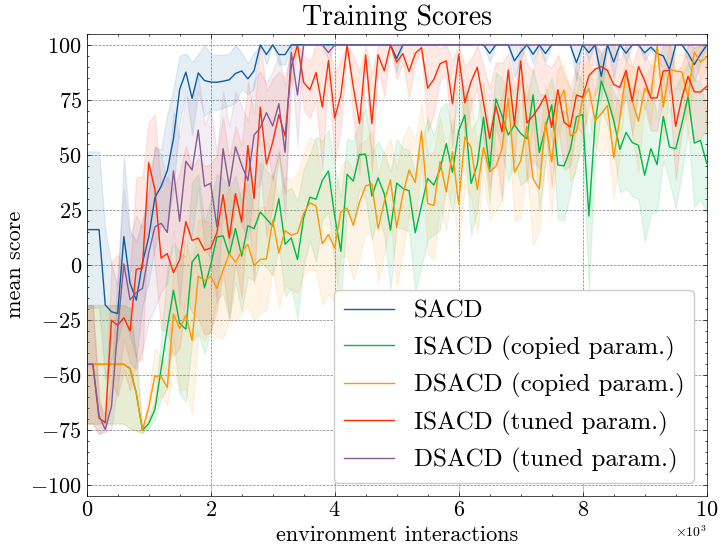}
    \caption{Training progress of the single-agent SACD, and two versions of multi-agents ISACD and DSACD, using the original and tuned parameters, respectively (cf.~\cref{Table:Params}).} \vspace{-0.25cm}
    \label{Fig:Plot_SACD_MASACD}
\end{figure}

\subsection{Hyperparameters}
Both PPO and SACD algorithms have multiple hyperparameters that influence the progress of training. For single-agent algorithms, we used default parameters both for PPO and SACD based on the literature \cite{schulman2017proximal,christodoulou2019soft,yoon2021winning}. However, as mentioned in \cite{yu2022surprising}, multi-agent RL may require different hyperparameters compared to single-agent RL for optimal performance. Therefore, in the MARL context, we tuned the hyperparameters using Optuna, an open-source hyperparameter optimization framework~\cite{akiba2019optuna}. \cref{Table:Params} summarizes the hyperparameters we used both in the single-agent and multi-agent settings.

\begin{table}[!tbh]
\begin{tabular}{l|llll}
\hline
Parameters                  & PPO    & MAPPO    & SACD     & MASACD \\ \hline
(Mini-)Batch size                  & 4 x 32 & 2 x 32   & 64       & 16     \\
Update start                &        &          & 4        & 3      \\
Discount ($\gamma$)         & 0.95   & 0.996    & 0.995    & 0.998  \\
Learning-rate               & 0.003  & 0.002    & $5 \times 10^{-5}$ & 0.0002 \\
VF coeff. ($c_1$)       & 0.5    & 0.5     &          &        \\
Entropy coeff. ($c_2$)             & 0.01   & $5 \times 10^{-5}$ &          &        \\
Clipping param. ($\epsilon$) & 0.2    & 0.12     &          &        \\
GAE param. ($\lambda$)               & 0.95   & 0.85     &          &        \\
Target entropy scale        &        &          & 0.98     & 0.98   \\
Tau                         &        &          & 0.001    & 0.002  \\ \hline
\end{tabular}
\caption{Parameters used for the different RL algorithms. The two multi-agent versions of PPO (IPPO and DPPO) use the same parameters, reported in the column MAPPO. Similarly, the column MASACD reports the parameters used for both the multi-agent versions of SACD (ISACD and DSACD).}
\label{Table:Params}
\end{table}

We did not perform any hyperparameter tuning for the design of the actor and critic networks. For training, we used the Adam optimizer. In both single- and multi-agent SACD, we used 3 GNN blocks in the shared layer with a dimension of 128. For the actor, we use 3 GNN blocks again, and for the critic 1 GNN block. Furthermore, in both the single- and multi-agent PPO we used 3 GNN blocks in both the critic and the actor.

\begin{figure}[!t]
    \centering
    \includegraphics[width=0.99\linewidth]{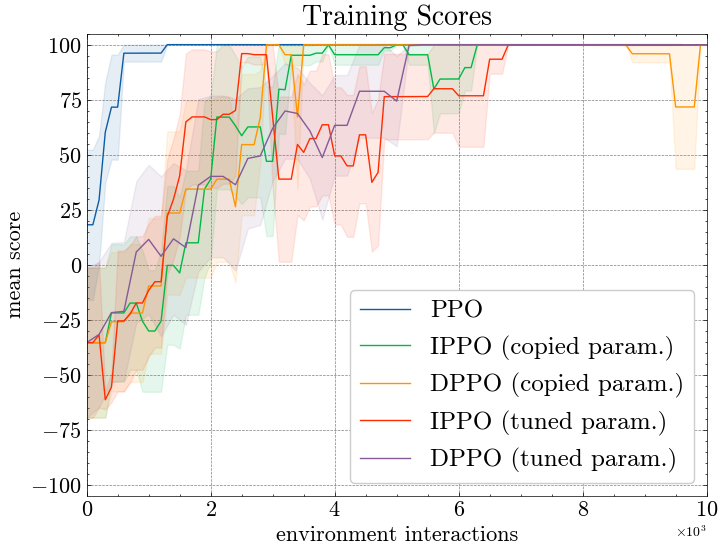}
    \caption{Training progress of the single-agent PPO, and two versions of multi-agents IPPO and DPPO, using the original and tuned parameters, respectively (cf.~\cref{Table:Params}).} \vspace{-0.25cm}
    \label{Fig:Plot_PPO_MAPPO}
\end{figure}

\subsection{Comparison of Single-Agent RL vs.~Multi-Agent RL}
In this section, we present the main results for single-agent and multi-agent RL architectures. In all plots, we show the mean episode score (solid line) and the corresponding standard error (shaded area) averaged over different model seeds to give insight into the training progress.

\subsubsection{SACD algorithm}
\cref{Fig:Plot_SACD_MASACD} shows the results of the single- and multi-agent versions of SACD. The single-agent version is able to achieve the optimal score but seems to remain unstable until the end. 
\cref{Fig:Plot_SACD_MASACD} also reports the scores of ISACD and DSACD deployed either (i) using hyperparameters identical to those used in the single-agent SACD (cf.~the SACD column in \cref{Table:Params}) or (ii) using ad-hoc hyperparameters obtained using Optuna. It is clear that hyperparameters optimized for the single-agent SACD version exhibit suboptimal performance when directly applied to their multi-agent counterparts. With optimized parameters, ISACD achieves peak scores but suffers from instability, performing significantly worse than SACD, as expected in~\cref{Sec:DSACD}. The DSACD agent with tuned parameters achieves an optimal score after a number of environment interactions comparable to those needed by the single agent. However, this agent is much more stable and, in fact, maintains a perfect optimal score until the training ends. 

\subsubsection{PPO algorithm}
\cref{Fig:Plot_PPO_MAPPO} illustrates the results for the single- and multi-agent versions of PPO. Even if the single-agent version converges more quickly, all agents are able to find the optimal solution. In terms of hyperparameters, the score difference between multi-agent algorithms with or without optimized parameters is less significant than when using SACD. This can be explained by the fact that the difference between the hyperparameters in PPO and MAPPO is modest and by the fact that PPO is less sensitive to hyperparameter changes. In the single-agent case, PPO outperforms SACD, but looking at the multi-agent setting, they achieve the maximum score after a similar number of environment interactions. Note that the independent version of PPO is not as unstable as ISACD.

\begin{figure}[!t]
    \centering
    \includegraphics[width=0.99\linewidth]{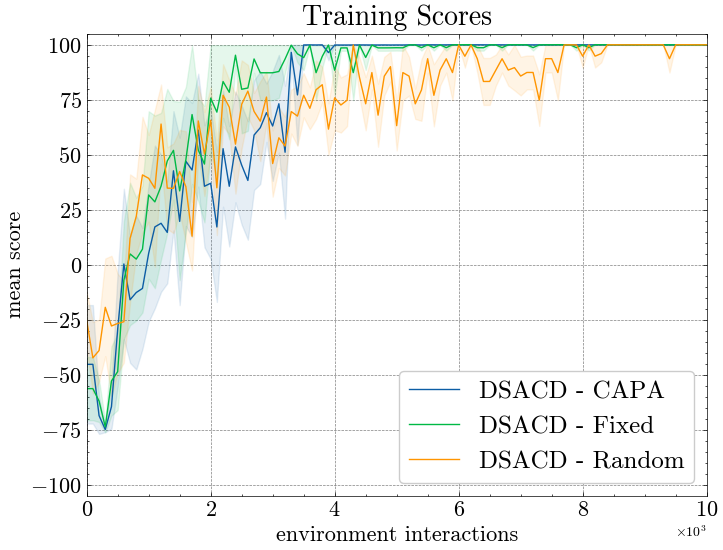}
    \caption{Training progress using different policies for the mid-level agent for DSACD.}
    \label{Fig:Plot_mid_DSACD} \vspace{-0.3cm}
\end{figure}

\subsection{Comparing different mid-level policies}
\begin{figure}[!tb]
    \centering
    \includegraphics[width=0.99\linewidth]{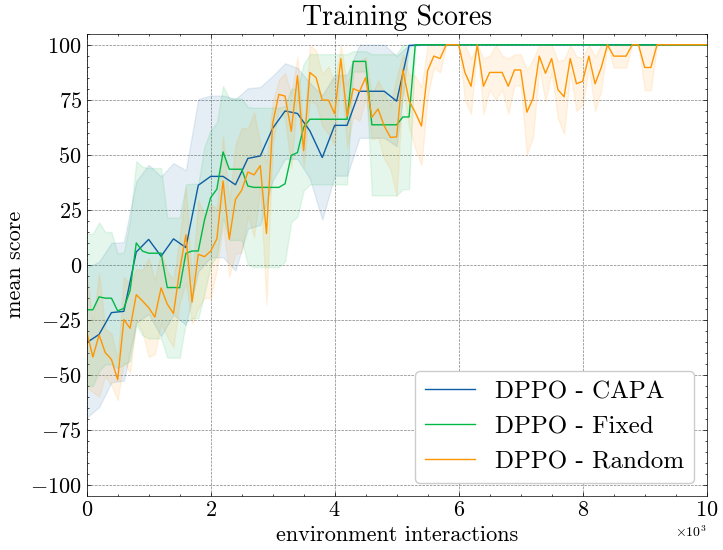}
    \caption{Training progress using different policies for the mid-level agent for DPPO.}
    \label{Fig:Plot_mid_DPPO} \vspace{-0.3cm}
\end{figure}

For completeness, we investigate the effects of using other policies for the mid-level agent, which are different from the CAPA policy described in \cref{Sec:MiddleAgent}. More specifically, we consider a \textit{fixed policy}, in which the mid-level agent orders the low-level agents based on the size of the corresponding substations, and a \textit{random policy}, in which every time the low-level agents act in random order. The results of these tests are shown in \cref{Fig:Plot_mid_DSACD,Fig:Plot_mid_DPPO}. The random policy results in more unstable behavior than the other two. 
The difference between CAPA and the fixed policy is more difficult to appreciate. 
It makes sense that with random patterns in the order of the agents, the agents have more trouble learning the right actions compared to a more fixed order. We expect that when training both the mid-level and lowest-level agents simultaneously this will become a challenge.

\section{Conclusion} \label{sec:conclusion}
This paper presents a hierarchical Multi-Agent Reinforcement Learning (MARL) framework for power network management through topological actions. Our architecture uses rule-based agents at the highest and mid levels, emphasizing varied MARL strategies for the lowest level. Future work will develop learning policies for these agents, addressing inter-level dependencies, and expand experiments to larger networks, further amplifying MARL's advantages over single-agent systems.

\bibliographystyle{elsarticle-num}
\bibliography{references}
\end{document}